\documentclass[10pt,twocolumn,letterpaper]{article}

\usepackage{iccv}
\usepackage{times}
\usepackage{epsfig}
\usepackage{graphicx}
\usepackage{amsmath}
\usepackage{amssymb}

\usepackage{multirow}
\usepackage{epsfig,epstopdf,subcaption}

\usepackage{capt-of,etoolbox}

\usepackage[pagebackref=true,breaklinks=true,letterpaper=true,colorlinks,bookmarks=false]{hyperref}

\iccvfinalcopy % *** Uncomment this line for the final submission

\def\iccvPaperID{0737} % *** Enter the ICCV Paper ID here

\ificcvfinal\fi

\begin{document}

%%%%%%%%% TITLE
\title{Multimodal Style Transfer via Graph Cuts}
\vspace{-8mm}
\author{Yulun Zhang$^{1,2}$, Chen Fang$^{2,3}$, Yilin Wang$^2$, Zhaowen Wang$^2$, Zhe Lin$^2$, Yun Fu$^1$, Jimei Yang$^2$\\
$^1$Northeastern University, $^2$Adobe Research, $^3$ByteDance AI Lab
%\author{Yulun Zhang$^1$, Chen Fang$^2$, Yilin Wang$^2$, Zhaowen Wang$^2$, Zhe Lin$^2$, Yun Fu$^1$, Jimei Yang$^2$\\
%$^1$Northeastern University, $^2$Adobe Research \\
%{\tt\small yulun100@gmail.com, fangchen@bytedance.com, \{ @adobe.com}
%Institution1 address %\\
%\and
}

\makeatletter
\let\@oldmaketitle\@maketitle
\renewcommand{\@maketitle}{\@oldmaketitle
\centering
\vspace{-8mm}
\includegraphics[width=160mm]{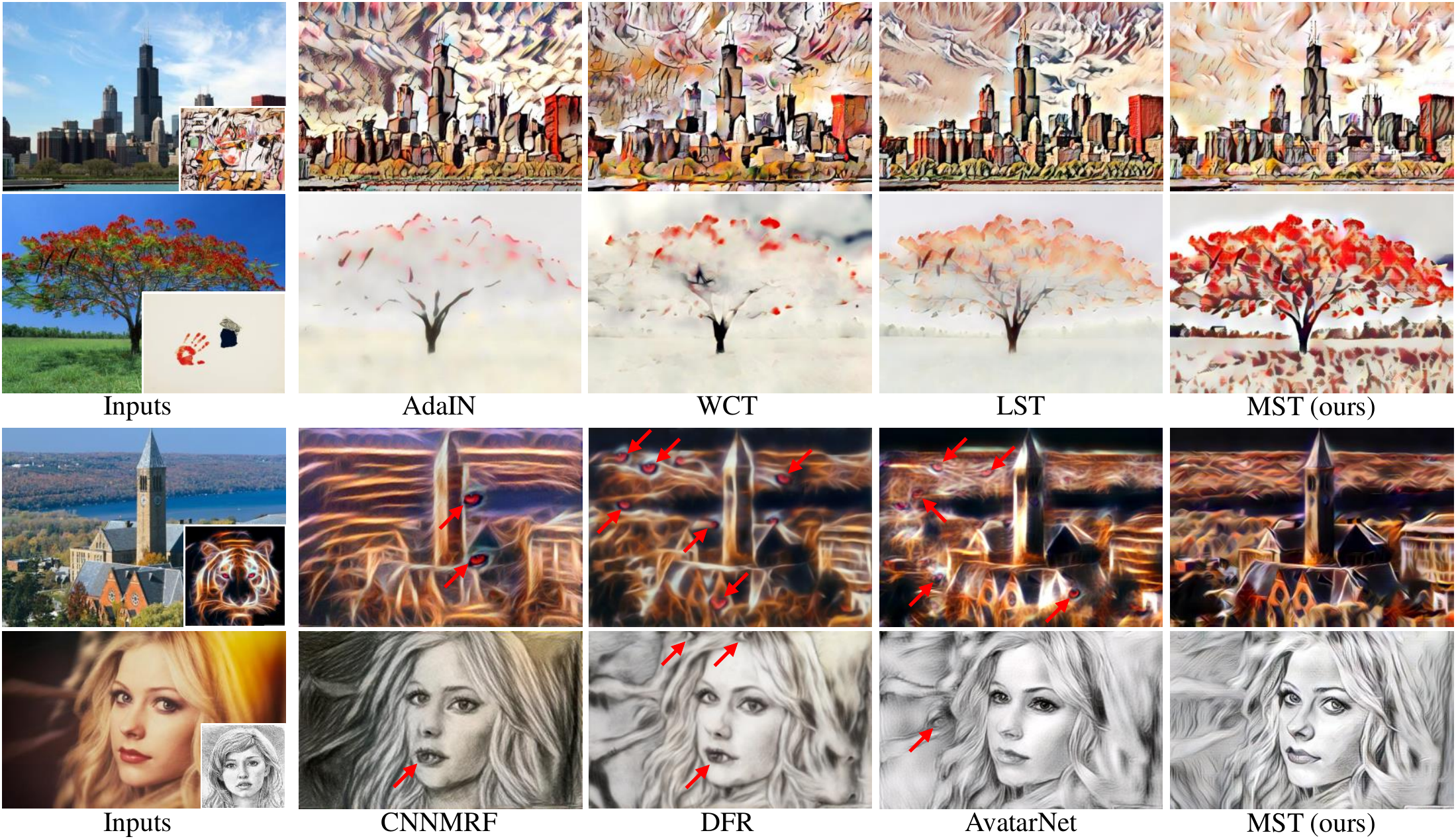} 
\vspace{-3mm}
\captionof{figure}{Gram matrix based style transfer methods (AdaIN~\cite{huang2017arbitrary}, WCT~\cite{li2017universal}, and LST~\cite{li2019learning}) may fail to distinguish style patterns (1st and 2nd rows). Patch-swap based methods (CNNMRF~\cite{li2016combining}, DFR~\cite{gu2018arbitrary}, and AvatarNet~\cite{sheng2018avatar}) may copy some less desired style patterns (labeled with {\color{red}red arrows}) to the results (3rd and 4th rows). Our MST alleviates all these limitations.}
\label{fig:fig_first}
\vspace{2mm}
}
\makeatother

\maketitle
\ificcvfinal\thispagestyle{empty}\fi

%%%%%%%%% ABSTRACT
\begin{abstract}
\vspace{-5mm}
An assumption widely used in recent neural style transfer methods is that image styles can be described by global statics of deep features like Gram or covariance matrices. Alternative approaches have represented styles by decomposing them into local pixel or neural patches. Despite the recent progress, most existing methods treat the semantic patterns of style image uniformly, resulting unpleasing results on complex styles. In this paper, we introduce a more flexible and general universal style transfer technique: multimodal style transfer (MST). MST explicitly considers the matching of semantic patterns in content and style images. Specifically, the style image features are clustered into sub-style components, which are matched with local content features under a graph cut formulation. A reconstruction network is trained to transfer each sub-style and render the final stylized result. We also generalize MST to improve some existing methods. Extensive experiments demonstrate the superior effectiveness, robustness, and flexibility of MST.
\end{abstract}

%%%%%%%%% BODY TEXT
\vspace{-4mm}
\section{Introduction}
\vspace{-4mm}
Image style transfer is the process of rendering a content image with characteristics of a style image. Usually, it would take a long time for a diligent artist to create a stylized image with particular style. Recently, it draws a lot of interests~\cite{gatys2016image,johnson2016perceptual,chen2016fast,wang2017multimodal,huang2017arbitrary,li2017universal,gu2018arbitrary,sheng2018avatar,shen2018neural,chen2018stereoscopic,zhang2018separating,li2019learning} since Gatys \textit{et al.}~\cite{gatys2016image} discovered that the correlations between convolutional features of deep networks can represent image styles, which would have been hard for traditional patch-based methods to deal with. These neural style transfer methods either use an iterative optimization scheme~\cite{gatys2016image} or feed-forward networks~\cite{johnson2016perceptual,chen2016fast,wang2017multimodal,huang2017arbitrary,li2017universal,sheng2018avatar,li2019learning} to synthesize the stylizations. Most of them are applicable for arbitrary style transfer with a pre-determined model. These universal style transfer methods~\cite{huang2017arbitrary,li2017universal,sheng2018avatar,li2019learning} inherently assume that the style can be represented by the global statistics of deep features such as gram matrix \cite{gatys2016image} and its approximates \cite{huang2017arbitrary,li2017universal}.  Although these neural style transfer methods can preserve the content well and match the overall style of the reference style images, they will also distort the local style patterns, resulting unpleasing visual artifacts.

Let's start with some examples in Fig.~\ref{fig:fig_first}. In the first row, where the style image consists of complex textures and strokes, these methods cannot tell them apart and neglect to match style patterns to content structures adaptively. This would introduce some less desired strokes in smooth content areas, e.g., the sky. 
In the second row, the style image has clear spatial patterns (e.g., large uniform background and blue/red hand). AdaIN, WCT, and LST  failed to maintain the content structures and suffered from wash-out artifacts. This is mainly because the unified style background occupies a large proportion in the style image, resulting its domination in the global statistics of style features. These observations indicate that it may not be sufficient to represent style features as at may not be sufficient to represent style features as a unimodal distribution such as a Gram or covariance matrix. An ideal style representation should respect to the spatially-distributed style patterns.

%Interestingly, another line of research represents a style by local style features~\cite{chen2016fast,li2016combining,gu2018arbitrary,sheng2018avatar}. Inherited from traditional patch-based methods, these neural patch-based algorithms could generate visually pleasing results, if content and style images have similar structures. However, their example matching usually uses greedy algorithm, which may introduce less desired style patterns to the outputs. We illustrate it with two examples in Fig.~\ref{fig:fig_first}, where the style images have salient patterns, e.g., eyes. These methods may copy them to the buildings and the landscape as well as the lip of the girl. Moreover, in the 4th row of Fig.~\ref{fig:fig_first}, these methods also suffer from shape distortion problem, e.g., the girl's appearance has changed. This phenomenon apparently limits the choice of style images for these methods. 
Inherited from traditional patch-based methods, these neural patch-based algorithms could generate visually pleasing results when content and style images have similar structures. However, the greedy example matching usually employed by these methods will introduce less desired style patterns to the outputs. This is illustrated by the bottom two examples in Fig.~\ref{fig:fig_first}, where some salient patterns in the style images, e.g., the eyes and lips, are improperly copied to the buildings and landscape. Moreover, the last row of Fig.~\ref{fig:fig_first} also illustrates shape distortion problem of these methods; e.g., the appearance for the girl has changed. This phenomenon apparently limits the choice of style images for these methods.  
%This is mainly caused by the patc

To address these issues, we propose the multimodal style transfer (MST), a more flexible and general style transfer method that seeks for a sweet spot between parametric (gram matrix based) and non-parametric (patch based) approaches. Specifically, instead of representing the style with a unimodal distribution, we propose a multimodal style representation with graph based style matching mechanism to adaptively match the style patterns to a content image.

\begin{figure}[tpb]
%\centering
%\includegraphics[scale=0.5]{network_RIR.pdf}
\centerline{
%\hspace{0.2mm}
\includegraphics[width = 82mm]{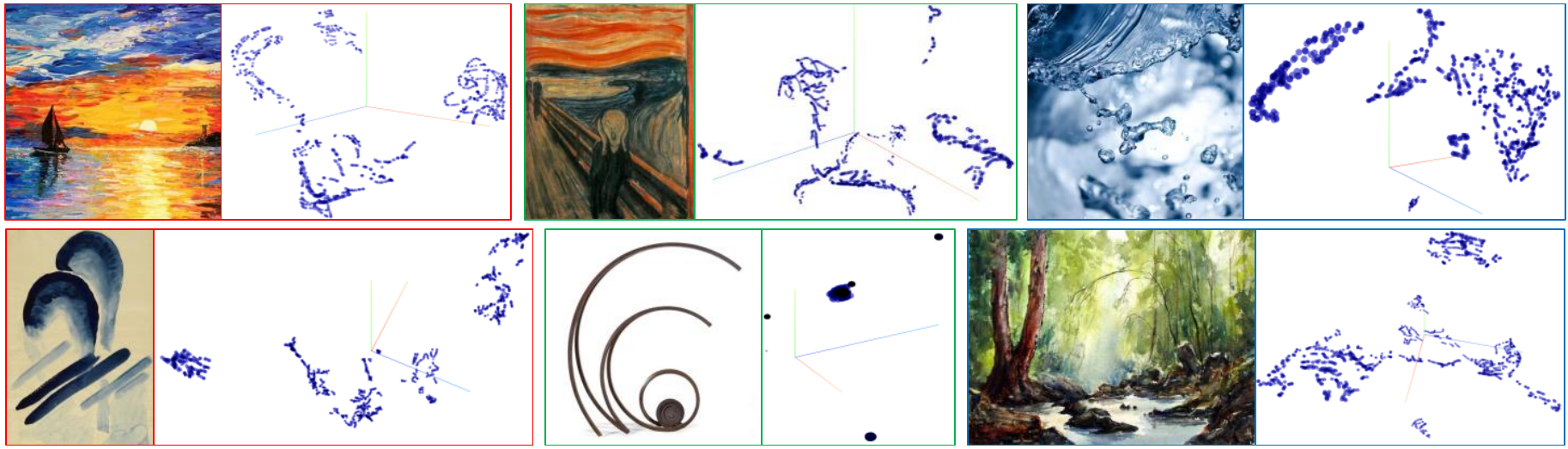}}
\vspace{-3mm}
\caption{t-SNE~\cite{maaten2008visualizing} visualization for style features. The original high-dimension style features are extracted at layer Conv\_4\_1 in VGG-19~\cite{simonyan2014very} and are reduced to 3 dimensions via t-SNE. We can see the feature distributions tend to fit as multimodal distributions rather than single-modal ones.}

\label{fig:style_tsne}
\vspace{-6mm}
\end{figure}

Our main contributions are summarized as follows:
\begin{itemize}
\item We analyze the feature distributions of different style images (see Fig.~\ref{fig:style_tsne}) and propose a multimodal style representation that better models the style feature distribution. This multimodal representation consists of a mixture of clusters, each of which represents a particular style pattern. It also allows users to mix-and-match different styles to render diverse stylized results.
\item We formulate style-content matching as an energy minimization problem with a graph and solve it via graph cuts. Style clusters are adapted to content features with respect to the content spatial configuration.   
\item We demonstrate the strength of MST by extensive comparison with several state-of-the-art style transfer methods. The robustness and flexibility of MST is shown with different sub-style numbers and multi-style mixtures. The general idea of MST can be extended to improve other existing stylization methods.
\end{itemize}

\vspace{-5mm}
\section{Related Works}
\vspace{-2mm}

\textbf{Style Transfer}. Originating from non-realistic rendering~\cite{kyprianidis2013state}, image style transfer is closely related to texture synthesis~\cite{efros1999texture,gatys2015texture,elad2017style}. Gatys \textit{et al.}~\cite{gatys2016image} were the first to formulate style transfer as the matching of multi-level deep features extracted from a pre-trained deep neural network, which has been widely used in various tasks~\cite{li2019rethinking,li2019gain,li2019vsr}. Lots of improvements have been proposed based on the works of Gatys \textit{et al.}~\cite{gatys2016image}. Johnson \textit{et al.}~\cite{johnson2016perceptual} trained feed-forward style-specific network and produced one stylization with one model. Sanakoyeu \textit{et al.}~\cite{sanakoyeu2018style} further proposed a style-aware content loss for high-resolution style transfer. Jing \textit{et al.}~\cite{jing2018stroke} proposed a StrokePyramid module to enable controllable stroke with adaptive receptive fields. However, these methods are either time consuming or have to re-train new models for new styles.

The first arbitrary style transfer was proposed by Chen and Schmidt~\cite{chen2016fast}, who matched each content patch to the most similar style patch and swapped them. Luan \textit{et al.}~\cite{luan2017deep} proposed deep photo style transfer by adding a regularization term to the optimization function. Based on markov random field (MRF), Li and Wand~\cite{li2016combining} proposed CNNMRF to enforce local patterns in deep feature space. Ruder~\textit{et al.}~\cite{ruder2016artistic} improved video stylization with temporal coherence. Although their visual stylizations for arbitrary style are appealing, the results are not stable~\cite{ruder2016artistic}. 

Recently, Huang \textit{et al.}~\cite{huang2017arbitrary} proposed real-time style transfer by matching the mean-variance statistics between content and style features. Li~\textit{et al.}~\cite{li2017universal} further introduced whitening and coloring (WCT) by matching the covariance matrices. Li \textit{et al.} boosted style transfer with linear style transfer (LST)~\cite{li2019learning}. Gu \textit{et al.}~\cite{gu2018arbitrary} proposed deep feature reshuffle (DFR), which connects both local and global style losses used in parametric and non-parametric methods. Sheng \textit{et al.}~\cite{sheng2018avatar} proposed AvatarNet to enable multi-scale transfer for arbitrary style. Shen \textit{et al.}~\cite{shen2018neural} built meta networks by taking style images as inputs and generating corresponding image transformation networks directly. Mechrez \textit{et al.}~\cite{mechrez2018contextual} proposed contextual loss for image transformation. However, these methods fail to treat the style patterns distinctively and neglect to adaptively match style patterns with content semantic information. For more neural style transfer works, readers can refer to the survey~\cite{jing2017neural}. 
%We provide a difference summary with others in supplementary material.  

\textbf{Graph Cuts based Matching}. Many problems that arose in early vision can be naturally expressed in terms of energy minimization. For example, a large number of computer vision problems attempt to assign labels to pixels based on noisy measurements. Graph cuts is a powerful method to solve such discrete optimization problems.  Greig \textit{et al.}~\cite{greig1989exact} were firstly successful solving graph cuts by using powerful min-cut/max-flow algorithms from combinatorial optimization. Roy and Cox~\cite{roy1998maximum} were the first to use these techniques for multi-camera stereo computation. Later, a growing number of researches in computer vision use graph-based energy minimization for a wide range of applications, which includes stereo~\cite{kolmogorov2002multi}, texture synthesis~\cite{kwatra2003graphcut}, image segmentation~\cite{veksler2000image}, object recognition~\cite{boykov1999new}, and others. In this paper, we formulate the matching between content and style features as an energy minimization problem. We approximate its global minimum via efficient graph cuts algorithms. To the best of our knowledge, we are the first to formulate style matching as energy minimization problem and solve it via graph cuts.
%More related works are available in .   
%However, it was almost 10 years later that the graph cut technique in~\cite{greig1989exact} was noticed, because   
%To the best of our knowledge, style transfer has not witnessed such a formulation for style matching.   
\vspace{-2mm}
\section{Proposed Method}
\label{sec:method}
\begin{figure}[t]
%\centering
%\includegraphics[scale=0.5]{network_RIR.pdf}
\centerline{
%\hspace{0.2mm}
\includegraphics[width=80mm]{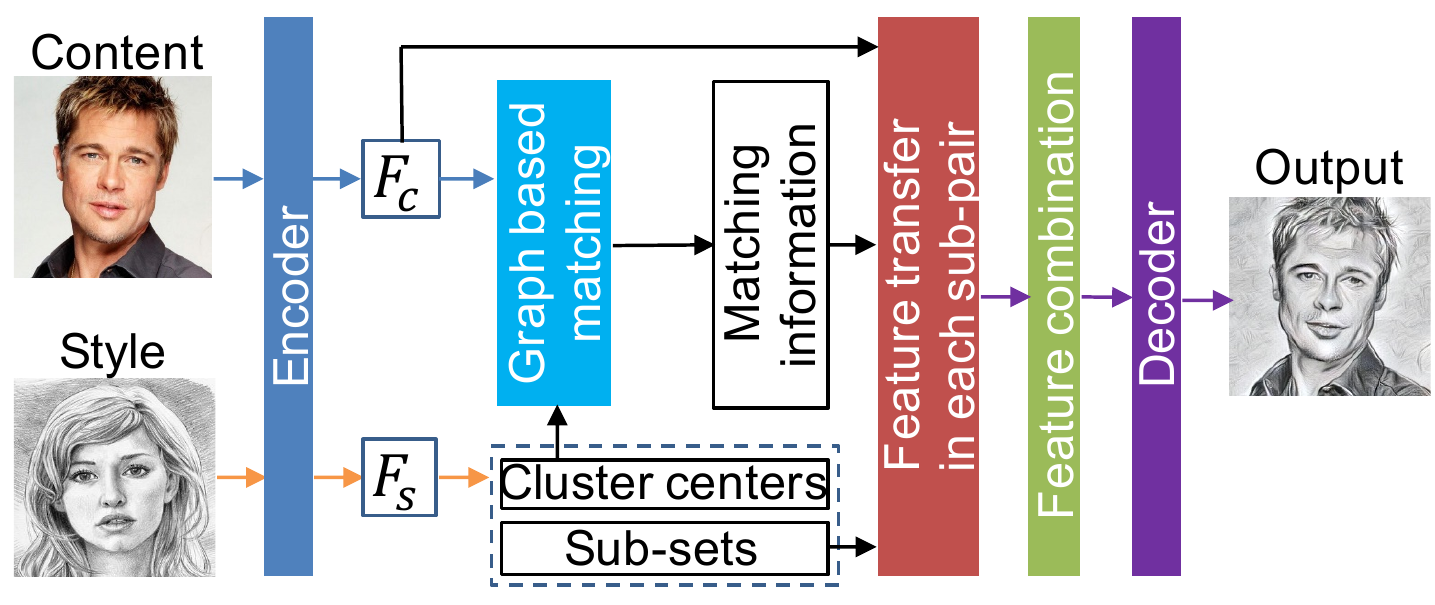}}
\vspace{-3mm}
\caption{An overview of our MST algorithm.}
\label{fig:pipeline_mst}
\vspace{-5mm}
\end{figure}

\vspace{-2mm}
We show the pipeline of our proposed MST in Fig.~\ref{fig:pipeline_mst}.
%We first investigate the style representation and propose a more efficient and reasonable multimodal style representation. Then we further show how to match each content feature with each sub-styles. Finally, we transform features in each sub-modal feature space. 

\vspace{-1mm}
\subsection{Multimodal Style Representation}
\label{subsec:multimodal_representation}
\vspace{-1mm}
In previous CNN-based image style transfer works, there are two main ways to represent style. One is to use the features from the whole image and assume that they are in the same distribution (e.g., AdaIN~\cite{huang2017arbitrary} and WCT~\cite{li2017universal}). The other one treats the style patterns as individual style patches (e.g., Deep Feature Reshuffle~\cite{gu2018arbitrary}).    
Equal treatments to different style patterns lack flexibility in the real cases, where there has several distributions among style features. Let's see the t-SNE~\cite{maaten2008visualizing} visualization for style features in Fig.~\ref{fig:style_tsne}, where the style features are clustered to multiple groups.  Therefore, if a cluster dominates the feature space, e.g.,  second example of Fig.~\ref{fig:fig_first}, the Gram matrix based methods \cite{li2017universal,li2019learning,huang2017arbitrary} will fail to capture the the overall style patterns. On the other hand, patch-based methods which treat each sub-patch distinctly would suffer from copying  multiple same style patterns to the results directly. For example, in Fig.~\ref{fig:fig_first}, the eyes in the style images are copied multiple times, causing unpleasing stylization results.  

Based on the observations and analyses above, we argue that neither a global statics of deep features nor local neural patches could be a suitable way to represent the complex real-world cases. As a result, we propose multimodal style representation, a more efficient and flexible way to represent different style patterns.

% % multimodal_style
\begin{figure}[t]
%\centering
%\includegraphics[scale=0.5]{network_RIR.pdf}
\centerline{
%\hspace{0.2mm}
\includegraphics[width=82mm]{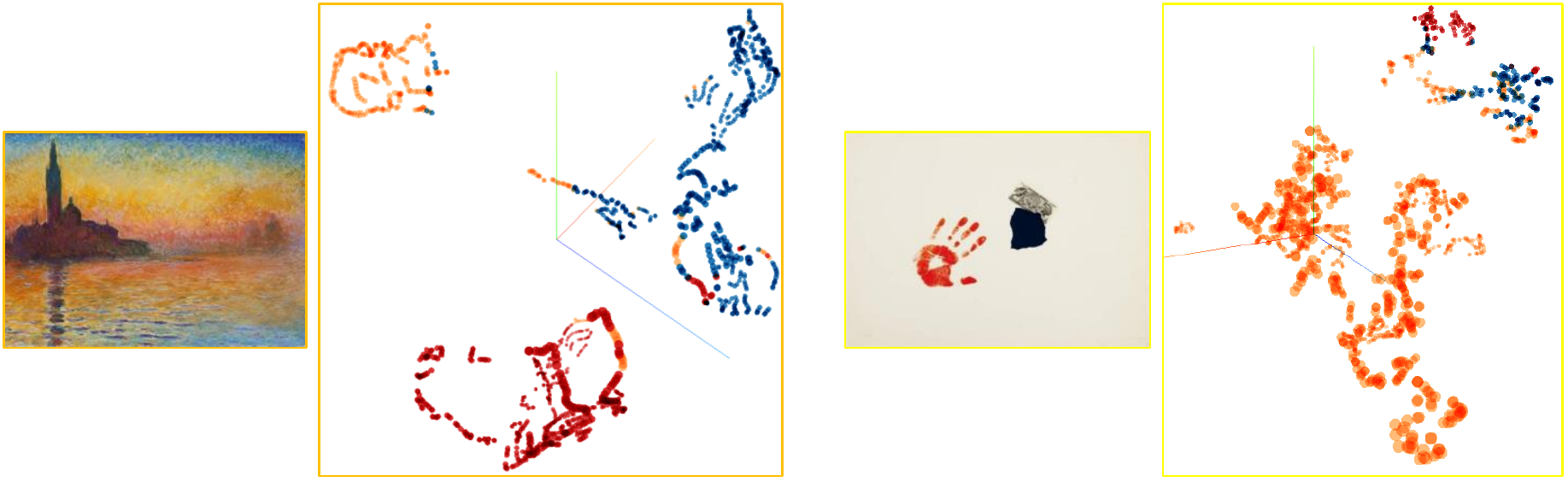}}
\vspace{-3mm}
\caption{t-SNE~\cite{maaten2008visualizing} visualization for style features with cluster labels. For each style-visualization pair, we set $K=3$ and label style features with corresponding cluster labels.}

\label{fig:multimodal_style}
\vspace{-5mm}
\end{figure}

%For a given style image $I_s$, we can extract its deep features $F_s\in \Re ^{C\times H_sW_s}$ via a pre-trained encoder $E_{\theta _{enc}}\left ( \cdot  \right )$, like VGG-19~\cite{simonyan2014very}. $H_s$ and $W_s$ are the height and width of the style feature.  To achieve multimodal representation in high-dimension feature space, we target to segment the style patterns into multiple subsets via Gaussian Mixture Model (GMM)~\cite{reynolds2015gaussian}. Technically, we simply apply K-means to cluster all the style feature points into $K$ clusters without considering spatial style information
For a given style image $I_s$, we can extract its deep features $F_s\in \Re ^{C\times H_sW_s}$ via a pre-trained encoder $E_{\theta _{enc}}\left ( \cdot  \right )$, like VGG-19~\cite{simonyan2014very}. $H_s$ and $W_s$ are the height and width of the style feature.  To achieve multimodal representation in high-dimension feature space, we target to segment the style patterns into multiple subsets. Technically, we simply apply K-means to cluster all the style feature points into $K$ clusters without considering spatial style information
\setlength{\abovedisplayskip}{2pt}
\setlength{\belowdisplayskip}{2pt}
\begin{align}
\begin{split}
F_{s}= F_{s}^{l_1}\cup F_{s}^{l_2} \cup \cdots \cup F_{s}^{l_k} \cup \cdots \cup F_{s}^{l_K},
\end{split}
\end{align} 
where $F_s^{l_k}\in \Re ^{C\times N_k}$ is the $k$-th cluster with $N_k$ features and we assign this cluster a label $l_k$. In the clustered space, features in the same cluster have similar visual properties and are likely drawn from the same distribution (resembling Gaussian Mixture Model~\cite{reynolds2015gaussian}). This process helps us obtain a multimodal representation of style. % for style features. 
%Here, to obtain multimodal representation, we simply apply K-means for clustering without considering spatial information.
%Then we try to match each content feature with a specific style cluster.
%$\sum_{k=1}^{K}N_{k}=H_{s}W_{s}$.    

We visualize multimodal style representation in Fig.~\ref{fig:multimodal_style}. For each style image, we extract its VGG feature (at layer Conv\_4\_1 in VGG-19) and cluster it into $K=3$ clusters. Then, we conduct t-SNE~\cite{maaten2008visualizing} visualization with the cluster labels. As shown in Fig.~\ref{fig:multimodal_style}, clustering results match our assumption of multimodal style representation well. Nearby feature points tend to be in the same cluster. These observation not only shows the multimodal style distribution, but also demonstrates that clustering is a proper way to model such a multimodal distribution.

\subsection{Graph Based Style Matching}
\vspace{-1mm}
%\textbf{Graph based style matching}. 
Like style feature extraction, we extract deep content features $F_c\in \Re ^{C\times H_cW_c}$ from a content image $I_c$. $H_c$ and $W_c$ are the height and width of the content feature. Distance measurement is the first step before matching. To reach a good distance metric, we should consider the scale difference between the content and style features. Computation complexity should also be taken into consideration, since all the content features will be used to match. Based on above analysis, we calculate the cosine distance between content feature $F_{c,p}\in \Re ^{C\times 1}$ and style cluster center $F_{s,l_k}\in \Re ^{C\times 1}$ as follows
\begin{align}
\begin{split}
\label{eq:consine_D}
D\left ( F_{c,p},F_{s,l_k} \right )=1-\frac{{F_{c,p}}^TF_{s,l_k}}{\left \| F_{c,p} \right \|\left \| F_{s,l_k} \right \|},
\end{split}
\end{align} 
where $\left ( \cdot \right )^{T}$ is transpose operation and $\left \| \cdot \right \|$ is magnitude of the feature vector. 

Then we target to find a labeling $f$ that assigns each content feature $F_{c,p}$ with a style cluster center label $f_{p}\in \left \{ l_1,l_2,\cdots ,l_K \right \}$. We formulate the disagreement between $f$ and content features as follows
\begin{align}
\begin{split}
\label{eq:E_data}
E_{data}\left ( f \right )=\sum_{p=1}^{HcWc}D\left ( F_{c,p}, F_{s,f_p} \right ),
\end{split}
\end{align}
where we name $E_{data}$ as data energy. Minimizing $E_{data}$ encourages $f$ to be consistent with content features. 

\begin{figure}[t]
%\centering
%\includegraphics[scale=0.5]{network_RIR.pdf}
\centerline{
%\hspace{0.2mm}
\includegraphics[width=65mm]{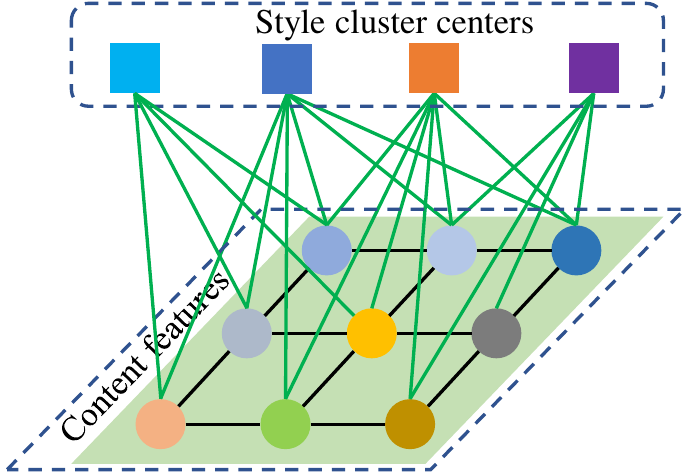}}
\vspace{-3mm}
\caption{Graph based style matching. Example of the graph containing content features and style cluster centers. We match content features with style cluster in pixel level.}

\label{fig:graph}
\vspace{-5mm}
\end{figure}

However, the spatial content information here is not considered, failing to preserve discontinuity and producing some unpleasing structures in the stylized results. Instead, we hope pixels in the same content local region have same labels. Namely, we want $f$ to be piecewise smooth and discontinuity preserving. So, we further introduce another smooth term $E_{smooth}\left ( f \right )$ as follows
\begin{align}
\begin{split}
\label{eq:E_s_v1}
E_{smooth}\left ( f \right )=\sum_{\left \{ p,q \right \}\in \Omega } V_{p,q}\left ( f_{p},f_{q} \right ),
\end{split}
\end{align}
where $\Omega$ is the position set of direct interacting pairs of content features. $V_{p,q}$ denotes the distinct penalty for each position pair of features $\left \{ p,q \right \}$. This has been investigated to be important in various computer vision applications~\cite{boykov2001fast}. Also, various forms of energy functions have been investigated before. Here, we take the discontinuity preserving function given by the Potts model
\begin{align}
\begin{split}
\label{eq:E_s_vpq}
V_{p,q}\left ( f_{p},f_{q} \right )=\lambda \cdot T\left ( f_{p}\neq f_{q} \right ),
\end{split}
\end{align}  
where $T\left ( \cdot  \right )$ is 1 if its argument is true, and otherwise 0. $\lambda$ is a smooth constant. This model encourages the labeling $f$ to pursue several regions, where content features in the same region have same style cluster labels.

By taking Eqs. \eqref{eq:E_data} and \eqref{eq:E_s_v1} into consideration, we naturally formulate the style matching problem as a minimization of the following energy function:
\begin{align}
\begin{split}
\label{eq:E_data_s}
E\left ( f \right )&=E_{data}\left ( f \right )+E_{smooth}\left ( f \right ). %\\
%&=\sum_{p=1}^{HcWc}D\left ( F_{c,p}, F_{s,f_p} \right )+ \sum_{\left \{ p,q \right \}\in \Omega } \lambda T\left ( f_{p}\neq f_{q} \right ).
\end{split}
\end{align} 
The whole energy $E\left ( f \right )$ measures not only the disagreement between $f$ and content features, but also the extent to which $f$ is not piecewise smooth. However, the global minimization of such an energy function is NP-hard even in the simplest discontinuity-preserving case~\cite{boykov2001fast}.

\begin{figure}[t]
%\centering
%\includegraphics[scale=0.5]{network_RIR.pdf}
\centerline{
%\hspace{0.2mm}
\includegraphics[width = 82mm]{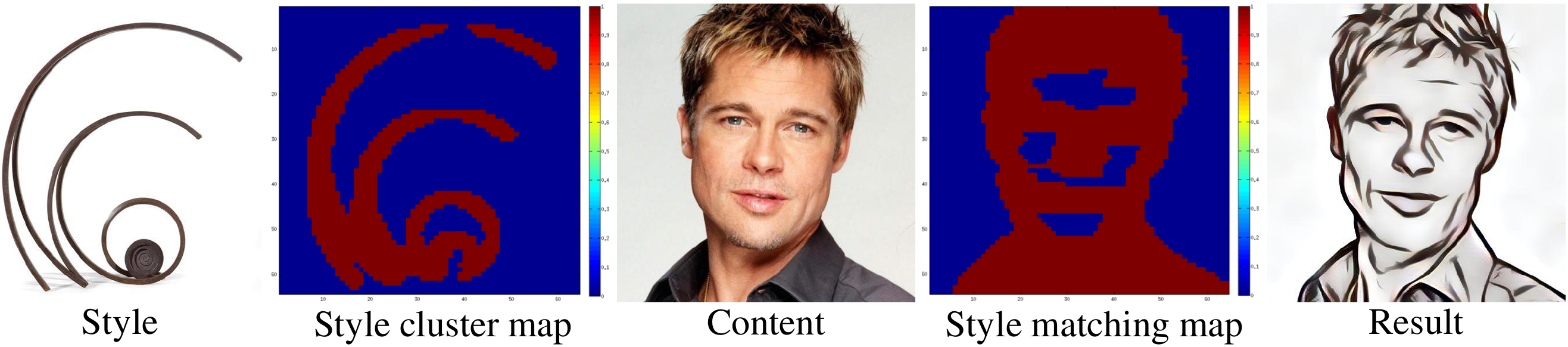}}
\vspace{-3mm}
\caption{Visualization of style matching. Here, we cluster style features into $K=2$ subsets for better understanding.}
\label{fig:style_match}
\vspace{-5mm}
\end{figure}

To solve the energy minimization problem in Eq.~\eqref{eq:E_data_s}, we propose to build a graph by regarding content features as $p$-vertices and style cluster centers as $l$-vertices (shown in Fig.~\ref{fig:graph}). Then the energy minimization is equal to min-cut/max-flow problem, which can be efficiently solved via graph cuts~\cite{boykov2001fast}. After finding a local minimum, the whole content features can be re-organized as follows    
\begin{align}
\begin{split}
F_{c}= F_{c}^{l_1}\cup F_{c}^{l_2} \cup \cdots \cup F_{c}^{l_k} \cup \cdots \cup F_{c}^{l_K},
\end{split}
\end{align}
where $F_{c}^{l_{k}}$ demotes the sub-set whose content features are matched with the same style label $l_{k}$. 

%\textbf{Style matching visualization}. 
We show visualization details about graph based style matching in Fig.~\ref{fig:style_match}. We extract style and content features from Conv\_4\_1 layer in VGG-19. Due to several downsampling modules in VGG-19, the spatial resolution of the features is much smaller than that of the inputs. We label the spacial style feature pixels with their corresponding cluster labels and obtain the style cluster maps. According to the style cluster maps in Fig.~\ref{fig:style_match}, we find that style feature clustering grasps semantic information from style images. 

After style matching in pixel level, we get the content-style matching map, which also reflects the semantic information, matching the content structures adaptively. Such an adaptive matching alleviates the wash-out artifacts, when the style is very simple or has large area of unified background. Then, we are able to conduct feature transform in each content-style pair group.

\subsection{Multimodal Style Transfer}
\vspace{-2mm}
For each content-style pair group $\left \{ F_{c}^{l_k}, F_{s}^{l_k} \right \}$, we first center them by subtracting their mean vectors $\mu \left ( F_{c}^{l_k} \right )$ and $\mu \left ( F_{s}^{l_k} \right )$ respectively. We conduct feature whitening and coloring as used in WCT~\cite{li2017universal}. 
\begin{align}
\begin{split}
F_{cs}^{l_k}=C_{s}W_{c}F_{c}^{l_k}+\mu \left ( F_{s}^{l_k} \right ),
\end{split}
\end{align}
where $W_{c}=E_{c}^{l_k}\left ( D_{c}^{l_k} \right )^{-\frac{1}{2}}\left ( E_{c}^{l_k} \right )^{T}$ is a whitening matrix and $C_{s}=E_{s}^{l_k}\left ( D_{s}^{l_k} \right )^{\frac{1}{2}}\left ( E_{s}^{l_k} \right )^{T}$ is a coloring matrix. $E_{c}^{l_k}$ and $D_{c}^{l_k}$ are diagonal matrix of eigenvalues and the orthogonal matrix of eigenvectors of the covariance matrix $F_{c}^{l_k}\left (  F_{c}^{l_k} \right )^{T}$. For style covariance matrix $F_{s}^{l_k}\left (  F_{s}^{l_k} \right )^{T}$, the corresponding matrices are $E_{s}^{l_k}$ and $D_{s}^{l_k}$. The reasons why we choose WCT to transfer features is its robustness and efficiency \cite{li2017universal,li2019learning}. More details about whitening and coloring are introduced in~\cite{li2017universal}.

After feature transformation, we may also want to blend transferred features with content features as did in previous works (e.g., AdaIN~\cite{huang2017arbitrary} and WCT~\cite{li2017universal}). Most previous works have to blend the whole transferred features with a unified content-style trade-off, which treats different content parts equally and is not flexible to the real-world cases. Instead, our multimodal style representation and matching make it possible to adaptively blend features. Namely, for each content-style pair group, we blend them via    
\begin{align}
\begin{split}
F_{cs}^{l_k}=\alpha_{k} F_{cs}^{l_k}+ \left ( 1-\alpha_{k}  \right )F_{c}^{l_k},
\end{split}
\end{align}
where $\alpha_{k}\in \left [ 0,1 \right ]$ is a content-style trade-off for specific labeled content features. After blending all the features, we obtain the whole transferred features
\begin{align}
\begin{split}
F_{cs}= F_{cs}^{l_1}\cup F_{cs}^{l_2} \cup \cdots \cup F_{cs}^{l_k} \cup \cdots \cup F_{cs}^{l_K}.
\end{split}
\end{align}
$F_{cs}$ is then fed into the decoder $D_{\theta _{dec}}\left ( \cdot  \right )$ to reconstruct the final output $I_{cs}$. 
%More details about the network training are shown below. 
%\vspace{-4mm}
\subsection{Implementation Details}
\vspace{-2mm}
%\footnote{The MST source code will be available after the paper is published.}
Now, we specify the implementation details about our proposed MST. Similar to some previous works (e.g., AdaIN, WCT, DFR), we incorporate the pre-trained VGG-19 (up to Conv\_4\_1)~\cite{simonyan2014very} as the encoder $E_{\theta _{enc}}\left ( \cdot  \right )$. We obtain decoder $D_{\theta _{dec}}\left ( \cdot  \right )$ by mirroring the encoder, whose pooling layers are replaced by nearest up-scaling layers.

To train the decoder, we use the pre-trained VGG-19~\cite{simonyan2014very} to compute perceptual loss $l_{total}=l_{c}+\gamma l_{s},$ which combines content loss $l_{c}$ and style loss $l_{s}$. We simply set the weighting constant as $\gamma=10^{-2}$. Inspired by the loss designations in~\cite{johnson2016perceptual,li2017demystifying,huang2017arbitrary}, we formulate content loss $l_{c}$ as 
\begin{align}
\begin{split}
\l _{c}
=\left \| \phi _{4\_1}\left ( I_{c} \right )-\phi _{4\_1}\left ( I_{cs} \right ) \right \|_{2},
\end{split}
\end{align}
where $\phi _{4\_1}\left ( \cdot \right )$ extracts features at layer Conv\_{$4$}\_1 in VGG-19. We then formulate the style loss $l_{s}$ as 
\begin{align}
\begin{split}
\l _{s}&=\sum_{i=1}^{4}\left ( \left \| \mu \left ( \phi _{i\_1}\left ( I_{s} \right ) \right )-\mu \left ( \phi _{i\_1}\left ( I_{cs} \right ) \right ) \right \|_{2}\right ) \\
&+ \sum_{i=1}^{4}\left ( \left \| \sigma  \left ( \phi _{i\_1}\left ( I_{s} \right ) \right )-\sigma  \left ( \phi _{i\_1}\left ( I_{cs} \right ) \right ) \right \|_{2} \right ),
\end{split}
\end{align}
where $\phi _{i\_1}\left ( \cdot \right )$ extracts features at layer Conv\_{$i$}\_1 in VGG-19. We use $\mu \left ( \cdot  \right )$ and $\sigma \left ( \cdot  \right )$ to compute the mean and standard deviation of the content and style features.

\begin{figure}[t]
%\centering
%\includegraphics[scale=0.5]{network_RIR.pdf}
\centerline{
%\hspace{0.2mm}
\includegraphics[width = 82mm]{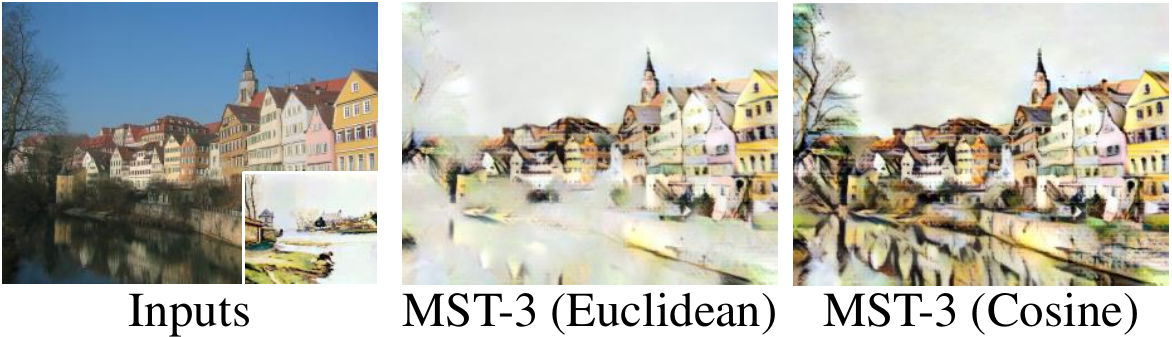}}
\vspace{-3mm}
\caption{Distance measurement investigation.}
\label{fig:distance}
\vspace{-3mm}
\end{figure}

\begin{figure}[tpb]
%\centering
%\includegraphics[scale=0.5]{network_RIR.pdf}
\centerline{
%\hspace{0.2mm}
\includegraphics[width = 82mm]{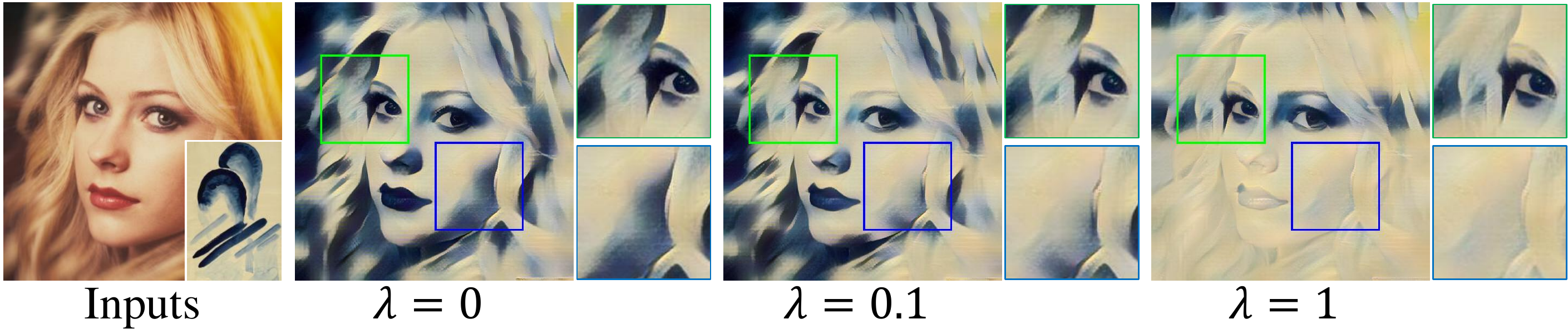}}
\vspace{-3mm}
\caption{Discontinuity preservation investigation.}
%Artifacts would be generated around the edges without smooth term ($\lambda=0$). Over emphasizing the smooth term (e.g., $\lambda=1$) would sacrifice the style diversity.
% A proper $\lambda$ is needed.
\label{fig:ablation_smooth}
\vspace{-5mm}
\end{figure}

We train our network by using images from MS-COCO~\cite{lin2014microsoft} and WikiArt~\cite{nichol2016painter} as content and style data respectively. Each dataset contains about 80,000 images. In each training batch, we randomly crop one pair of content and style images with the size of $256\times256$ as input. We implement our model with TensorFlow and apply Adam optimizer~\cite{kingma2014adam} with learning rate of $10^{-4}$.

\vspace{-2mm}
\section{Discussions}
\label{sec:discussions}
\vspace{-2mm}
%Despite the main differences (e.g., multimodal representation, graph based style matching, and multimodal style transfer) between MST and previous style transfer methods, we show more differences with some related works. 
To better position MST among the whole body of style transfer works, we further discuss and clarify the relationship between MST and some representative works. 

\textbf{Differences to CNNMRF}. CNNMRF~\cite{li2016combining} extracts a pool of neural patches from style images, with which patch matching is used to match content. MST clusters style features into multiple sub-sets and matches style cluster centers with content feature points via graph cuts. CNNMRF uses smoothness prior for reconstruction, while MST uses it for style matching only. CNNMRF minimizes energy function to synthesize the results. MST generates stylization results with a decoder. 

\textbf{Differences to MT-Net}. Both color and luminance are treated as a mixture of modalities in MT-Net~\cite{wang2017multimodal}. MST obtains multimodal representation from style features via clustering. It should also be noted that MT-Net has to train new models for new style images. While, MST is designed for arbitrary style transfer with a single model.

\textbf{Differences to WCT}. In WCT~\cite{li2017universal}, the decoder is trained by using only content data and loss. MST introduces additional style images for training. WCT uses multiple layers of VGG features and conducts multi-level coarse-to-fine stylization, which costs much more time and sometimes distorts structures. While, MST only transfers single-level content and style features. Consequently, even we set $K=1$ in MST, we achieve more efficient stylizations.

\vspace{-3mm}
\section{Experiments}
\vspace{-2mm}
%We provide more results in supplementary material.
We conduct extensive experiments to validate the contributions of each component in our method, the effectiveness of our method, and the flexibility for user control.% More results are provided in supplementary material.

%\vspace{-2mm}
\subsection{Ablation Study}
\label{subsec:ablation_study}
%\vspace{-2mm}

\textbf{Distance Measurement}. We first investigate the choice of the distance measurement, as it is critical for graph building. Here, we mainly investigate Euclidean distance and cosine distance (shown in Eq.~\eqref{eq:consine_D}). As shown in Fig.~\ref{fig:distance}, MST with Euclidean distance is affected by the huge background and may fail to transfer desired style patterns, leading to wash-out artifacts. This is mainly because there is no normalization for the deep features. As a result, the weight of style cluster center is proportional to its spatial proportion, weakening its semantic meaning. Instead, MST with cosine distance performs much better. 
%This failure can hardly be retrieved by increasing $K$.
%Such simple yet efficient distance measurement also contributes to robustness of MST with different $K$, which will be detailed in Section~\ref{subsec:style_cluster_number}.    

\textbf{Discontinuity Preservation}. In Fig.~\ref{fig:ablation_smooth}, we show the effectiveness of the smooth term $E_{smooth}(f)$ in Eq.~\eqref{eq:E_data_s}. Specifically, we set $\lambda$ as 0, 0.1, and 1 respectively. In real-world style transfer, people would like to smooth the facial area, as they do in the real photos. Here, we select one portrait to investigate how $\lambda$ affects smoothness. When we set $\lambda=0$, the energy function in Eq.~\eqref{eq:E_data_s} is minimized by only considering the data term. This would introduce some unpleasing artifacts in the facial area near edges and demonstrate the necessity of smooth term. However, large smooth term (e.g., $\lambda=1$) would over-smooth the stylization results, decreasing the style diversity. A proper value of $\lambda$ would not only keep better smoothness, but also preserve style diversity. We empirically set $\lambda=0.1$ through the whole experiments. 
%As a result, we empirically set $\lambda=0.1$ through the whole experiments.         

 \begin{figure}[tpb]
 %\centering
 %\includegraphics[scale=0.5]{network_RIR.pdf}
 \centerline{
 %\hspace{0.2mm}
 \includegraphics[width = 82mm]{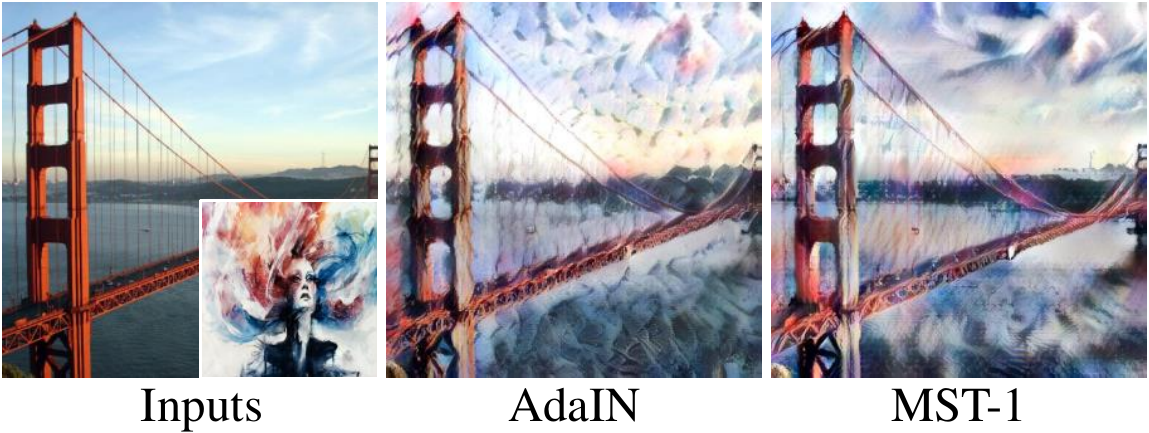}}
 \vspace{-3mm}
 \caption{Feature transformation investigation.}
 %Transferring features with WCT would alleviate the artifacts in the background and produce more natural stylizations than those with AdaIN.
 \label{fig:feature_transfer}
 \vspace{-6mm}
 \end{figure}

\textbf{Feature Transform}. Here we set $K=1$ in MST and compare with AdaIN~\cite{huang2017arbitrary} to show the effectiveness of WCT for feature transfer. As shown in Fig.~\ref{fig:feature_transfer}, AdaIN produces some stroke artifacts in the smooth area. These artifacts may make the cloud unnatural. This is mainly because AdaIN uses the mean and variance of the whole content/style features. Instead, by using whitening and coloring in a more optimized way, our MST-1 achieves more natural stylization and cleaner smoothed area. As a result, we introduce whitening and coloring for feature transform. 

\begin{figure*}[htbp]
%\centering
%\includegraphics[scale=0.5]{network_RIR.pdf}
\centerline{
%\hspace{0.2mm}
\includegraphics[width = 173mm]{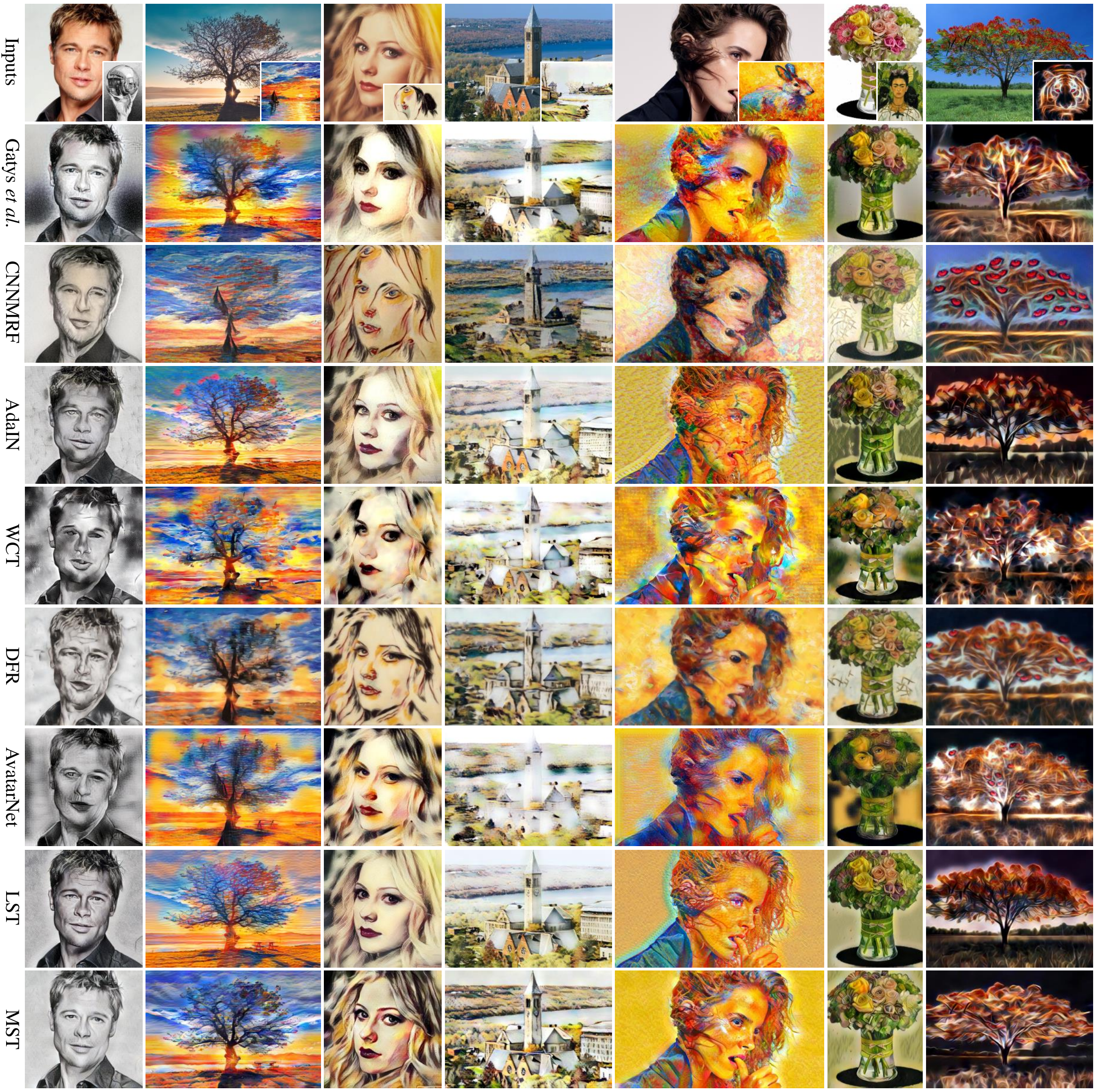}}
\vspace{-3mm}
\caption{Visual comparison. MST ($K=3$) and all compared methods use default parameters.}
% All the compared methods suffer from wash-out artifacts (e.g., 4th column), when the style contains large area of unified background.
\label{fig:visual_comp}
\vspace{-5mm}
\end{figure*}

\vspace{-2mm}
\subsection{Comparisons with Prior Arts}
\vspace{-2mm}
After investigating the effects of each component in our method, we turn to validate the effectiveness of our proposed MST. We compare with 7 state-of-the-art methods: method by Gatys \textit{et al.}~\cite{gatys2016image}, CNNMRF~\cite{li2016combining}, AdaIN~\cite{huang2017arbitrary}, WCT~\cite{li2017universal}, Deep feature reshuffle (DFR)~\cite{gu2018arbitrary}, AvatarNet~\cite{sheng2018avatar}, and LST~\cite{li2019learning}. We obtain results using their official codes and default parameters, except for Gatys \textit{et al.}\footnote{We use code from \href{https://github.com/jcjohnson/neural-style}{https://github.com/jcjohnson/neural-style}}.
%\footnote{We use code from \href{https://github.com/jcjohnson/neural-style}{https://github.com/jcjohnson/neural-style} with its default parameters (e.g., iterations = 10$^3$, learning rate = 1).}.

%After investigating the effects of each component in our method, we turn to validate the effectiveness of our proposed MST. We compare with 5 state-of-the-art methods: method by Gatys \textit{et al.}~\cite{gatys2016image}, AdaIN~\cite{huang2017arbitrary}, WCT~\cite{li2017universal}, Deep feature reshuffle (DFR)~\cite{gu2018arbitrary}, and AvatarNet~\cite{sheng2018avatar}. We obtain the compared results using their public released codes and default parameters: Gatys \textit{et al.}\footnote{\href{https://github.com/jcjohnson/neural-style}{https://github.com/jcjohnson/neural-style}} (iterations=10$^3$, learning rate=1), AdaIN\footnote{\href{https://github.com/xunhuang1995/AdaIN-stylee}{https://github.com/xunhuang1995/AdaIN-style}}, WCT\footnote{\href{https://github.com/Yijunmaverick/UniversalStyleTransfer}{https://github.com/Yijunmaverick/UniversalStyleTransfer}}, DFR\footnote{\href{https://github.com/msracver/Style-Feature-Reshuffle}{https://github.com/msracver/Style-Feature-Reshuffle}}, and AvatarNet\footnote{\href{https://github.com/LucasSheng/avatar-net}{https://github.com/LucasSheng/avatar-net}}.

%\href{https://github.com/jcjohnson/neural-style}{https://github.com/jcjohnson/neural-style}

\textbf{Qualitative Comparisons}. We show extensive comparisons~\footnote{The fifth content is from \href{https://www.mordeo.org/wallpapers/emma-watson-close-click}{https://www.mordeo.org}} in Fig.~\ref{fig:visual_comp}. Gatys \textit{et al.}~\cite{gatys2016image} transfer style with iterative optimization, being likely to falling in local minimum (e.g., 1st and 3rd columns). AdaIN~\cite{huang2017arbitrary} often produces less desired artifacts in the smooth area and some halation around the edges (e.g., 1st, 5th, and 6th columns). CNNMRF~\cite{li2016combining} may suffer from distortion effects and not preserve content structure well. Due to the usage of higher-level deep feature (e.g., Conv\_5\_1), WCT~\cite{li2017universal} would generate distorted results, failing to preserve the main content structures (e.g., 1st and 2nd columns). DRF~\cite{gu2018arbitrary} reconstructs the results by using the style patches, which could also distort the content structure (e.g., 1st and 3rd columns). In some cases (e.g., 5th, 6th, and 7th columns), some tiny style patterns (e.g., the eyes in the flowers and tree) would be copied to the results, leading unpleasing stylizations. AvatarNet~\cite{sheng2018avatar} would introduce some less desired style patterns in the smooth area (e.g., 1st column) and also copy some style patterns in the results (e.g., 6th and 7th columns). LST~\cite{li2019learning} could generate very good result in some cases (e.g., 6th column). However, it may suffer from wash-out artifacts (e.g., 3rd and 4th columns) and halation around the edges (e.g., 5th column). These compared methods mainly treat the style patterns as a whole, lacking distinctive ability to style patterns.   

%Except for the above discussed reasons causing the compared methods unpleasing stylizations, there is a more important factor. Namely, all the compared methods fail to treat style patterns distinctively. That's why all the compared methods would suffer from wash-out artifacts (e.g., 4th column), when the style image contains large part of unified background. The compared methods also neglect to adaptively conduct style transfer according the semantic content information, resulting in adding unwanted artifacts to the background of the portrait (e.g., 1st and 5th columns).    

Instead, we treat style features as multimodal presentations in high-dimension space. We match each content feature to its most related style cluster and adaptively transfer features according to the content semantic information. These advantages help explain why MST generates clearer results (e.g., 1st, 3rd, 5th, and 7th columns), performs more semantic matching with style patterns (e.g., 2nd column), and alleviates wash-out artifacts (e.g., 4th column). Such superior results demonstrate the effectiveness of our MST. 

\begin{table}[htbp]
%\scriptsize
\footnotesize
%\small
%\normalsize
\centering
\begin{center}
%\caption{Ablation investigation of contiguous memory (CM), local residual learning (LRL), and global feature fusion (GFF). We observe the best performance (PSNR) on Set5 with scaling factor $\times2$ in 200 epochs.} 
\vspace{-1mm}
\caption{Percentage of the votes that each method received.}
\label{tab:results_user_study} 
\vspace{-3mm}
%\begin{tabular*}{82.4mm}{@{\extracolsep{-0.75mm}}|c|c|c|c|c|c|c|c|c|}
\begin{tabular}{|c|c|c|c|c|c|c|c|c|c|c|c|}
\hline
Method & Gatys  & AdaIN  & WCT & DFR & AvatarNet & MST 
\\
\hline
Perc./\%& 21.41  & 11.31  & 12.67 & 11.55 & 9.61 & \textbf{33.45} 
\\
\hline
\end{tabular}
\end{center}
\vspace{-6mm}
\end{table}

\textbf{User Study}.
To further evaluate the 6 methods shown in Fig~\ref{fig:visual_comp}, we conduct a user study like~\cite{li2017universal}. We use 15 content images and 30 style images. For each method, we use the released codes and default parameters to generate 450 results. 20 content-style pairs are randomly selected for each user. For each style-content pair, we display the stylized results of 6 methods on a web-page in random order. Each user is asked to vote the one that he/she like the most. Finally, we collect 2,000 votes from 100 users and calculate the percentage of votes that each method received. The results are shown in Tab.~\ref{tab:results_user_study}, where our MST ($K$=3) obtains $33.45\%$ of the total votes. It's much higher than that of Gatys \textit{et al.}~\cite{gatys2016image}, whose stylization results are usually thought to be high-quality. This user study result is consistent with the visual comparisons (in Fig.~\ref{fig:visual_comp}) and further demonstrate the superior performance of our MST.      
%\vspace{-5mm}

\begin{table}[htbp]
%\scriptsize
\footnotesize
%\small
%\normalsize
\centering
\begin{center}
%\caption{Ablation investigation of contiguous memory (CM), local residual learning (LRL), and global feature fusion (GFF). We observe the best performance (PSNR) on Set5 with scaling factor $\times2$ in 200 epochs.} 
%\vspace{-3mm}
%\caption{Running time (s) comparisons for $512\times512$ images. MST-$K$ means $K$ style clusters.}
\caption{Running time (s) comparisons.}
\label{tab:results_time} 
\vspace{-3mm}
%\begin{tabular*}{82.4mm}{@{\extracolsep{-0.75mm}}|c|c|c|c|c|c|c|c|c|}
\begin{tabular}{|c|c|c|c|c|c|c|c|c|c|c|c|}
\hline
Method & Gatys  & AdaIN  & WCT & DFR & AvatarNet  
\\
\hline
Time (s)& 116.46  & 0.09  & 0.92 & 54.32 & 0.33  
\\
\hline
Method & MST-1 & MST-2 & MST-3 & MST-4 & MST-5
\\
\hline
Time (s)& 0.20 & 1.10 & 1.40 & 1.97 & 2.27  
\\
\hline
\end{tabular}
\end{center}
\vspace{-8mm}
\end{table}

\textbf{Efficiency}.
We further compare the running time of our methods with previous ones~\cite{gatys2016image,huang2017arbitrary,li2017universal,gu2018arbitrary,sheng2018avatar}. Tab.~\ref{tab:results_time} gives the average time of each method on 100 image pairs with size of $512\times512$. All the methods are tested on a PC with an Intel i7-6850K 3.6 GHz CPU and a Titan Xp GPU. Our MST with different $K$ performs relatively faster than methods by Gatys \textit{et al.}~\cite{gatys2016image} and DFR~\cite{gu2018arbitrary}. Even using SVD in CPU, MST-$1$ is faster than AvatarNet~\cite{sheng2018avatar} and WCT~\cite{li2017universal}. It should be noted that WCT conducts multi-level stylization, which costs much more time than that of MST-$1$. MST-$K$ ($K>1$) becomes much slower with larger $K$. This is mainly because our cluster operation is executed in CPU and consumes much more time. On the other hand, although MST with larger $K$ would consume more time, its stylized results would be very robust. So, in general, we don't have to choose very large $K$, of which we give more details about the effects later.     

%We also compare the running time of our methods with previous ones~\cite{gatys2016image,huang2017arbitrary,li2017universal,gu2018arbitrary,sheng2018avatar}. Tab.~\ref{tab:results_time} gives the average time of each method on 100 image pairs with size of $512\times512$. All the methods are tested on a PC with an Intel i7-6850K 3.6 GHz CPU and a Titan Xp GPU. Our MST with different $K$ performs relatively faster than methods by Gatys \textit{et al.}, DFR~\cite{gu2018arbitrary}, and WCT~\cite{li2017universal}. Even using SVD in CPU, MST-$1$ is faster than AvatarNet~\cite{sheng2018avatar}. While, MST-$K$ ($K>1$) is much slower. This is mainly because our cluster operation is executed in CPU and consumes much more time. More efficient GPU implementation of MST would be included in our future works. On the other hand, although MST with larger $K$ would consume more time, its stylized results would be very robust. So, in general, we don't have to choose very large $K$, of which we give more details about the effects later.  

\begin{figure*}[t]
%\centering
%\includegraphics[scale=0.5]{network_RIR.pdf}
\centerline{
%\hspace{0.2mm}
%width = 170mm
\includegraphics[width = 160mm]{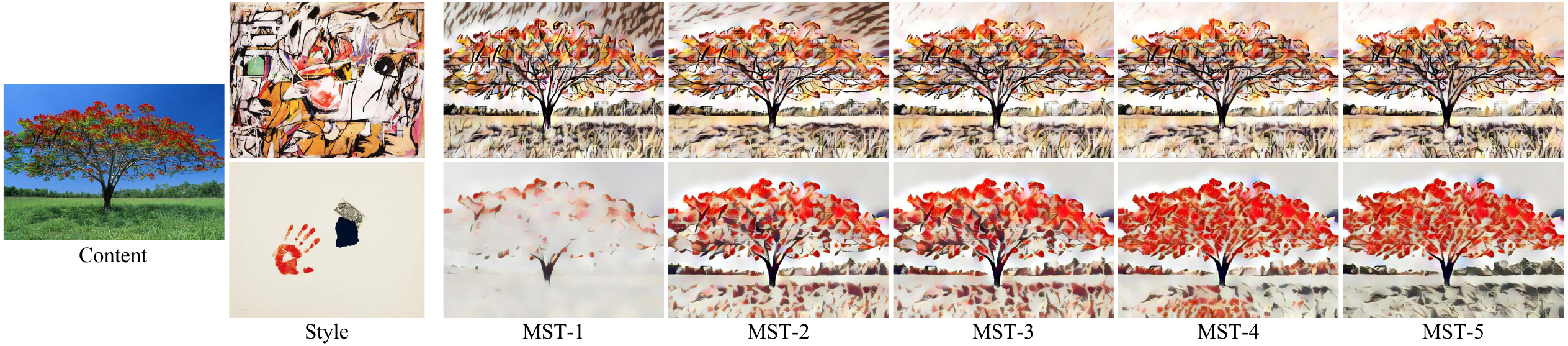}}
\vspace{-3mm}
\caption{Style cluster number investigation. Same content image with complex and simple style images.}

\label{fig:StyleClusterNumber}
\vspace{-3mm}
\end{figure*}

\begin{figure*}[t]
%\centering
%\includegraphics[scale=0.5]{network_RIR.pdf}
\centerline{
%\hspace{0.2mm}
\includegraphics[width = 160mm]{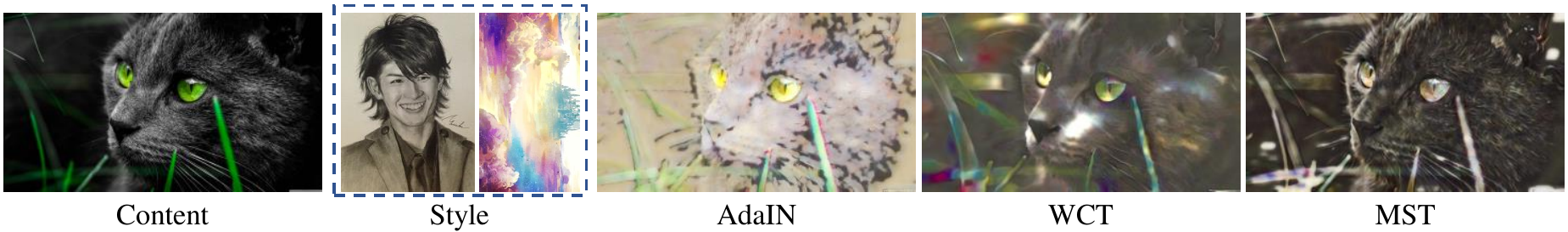}}
\vspace{-3mm}
\caption{Multi-style transfer. MST treats patterns from different style images distinctively and transfers them adaptively.}

\label{fig:multistyle}
\vspace{-5mm}
\end{figure*}

\subsection{Style Cluster Number} 
\label{subsec:style_cluster_number}
\vspace{-2mm}
We investigate how style cluster number $K$ affects the stylization in Fig.~\ref{fig:StyleClusterNumber}. When $K=1$, our MST performs style transfer by taking the whole style features equally, resulting in either very complex (1st row) or simple (2nd row) stylizations. These results are not consistent with the content structures and lack flexibility, leading to unpleasing feelings to users. Instead, we can produce multiple results with different $K$. When we enlarge $K$ with multimodal style representation, stylization results would either throw unnecessary style patterns (1st row) or introduce more matched style patterns (2nd row). The stylizations become more matched with the content structures. This is mainly because multimodal style representation allows distinctive and adaptive treatment for the style patterns. More important, MST reconstructs several stylization results with different $K$, providing multiple selections for the users.   
%We investigate how style cluster number $K$ affects the stylization in Fig.~\ref{fig:StyleClusterNumber}. When $K=1$, our MST performs style transfer by taking the whole style features equally, resulting in either very complex (1st row) or simple (2nd row) stylizations. These results are not consistent with the content structures and lack flexibility, leading to unpleasing feelings to users. Such cases are neglected by previous style transfer methods. Instead, we can produce multiple results with different $K$. When we enlarge $K$ with multimodal style representation, stylization results would either throw unnecessary style patterns (1st row) or introduce more matched style patterns (2nd row). The stylizations become more matched with the content structures. This is mainly because multimodal style representation allows distinctive and adaptive treatment for the style patterns. More important, MST reconstructs several stylization results with different $K$, providing multiple selections for the users.           

%\textbf{Content-Style Trade-off}. 
\vspace{-2mm}
\subsection{Adaptive Multi-Style Transfer}
\vspace{-2mm}
Most previous style transfer methods enable style interpolation, which blends the content image with a set of weighted stylizations. However, we don't fix the weights for each style image, but adaptively interpolate the style patterns to the content. As shown in Fig.~\ref{fig:multistyle}, the content image~\footnote{It's from \href{https://wallpaperstream.com/collection/cat/Green-Eye-Black-Cat-Wallpaper}{https://wallpaperstream.com}} is stylized by two style images simultaneously. We use AdaIN~\cite{huang2017arbitrary} and WCT~\cite{li2017universal} for reference (because it's not strictly fair comparisons) by setting equal weight for each style image. In Fig.~\ref{fig:multistyle}, AdaIN and WCT suffer from wash-out artifacts. While, our MST preserves the content structures well. MST transfers more portrait hair style to the cat body and more cloud style to the cat eyes and green leaves. Our adaptive multi-style transfer is also similar to spatial control in previous methods~\cite{huang2017arbitrary,li2017universal}. But, they need additional manually designed mask as input, consuming more user efforts. Instead, MST automatically allows good matching between content and style features.

\begin{figure}[tpb]
%\centering
%\includegraphics[scale=0.5]{network_RIR.pdf}
\centerline{
%\hspace{0.2mm}
\includegraphics[width = 82mm]{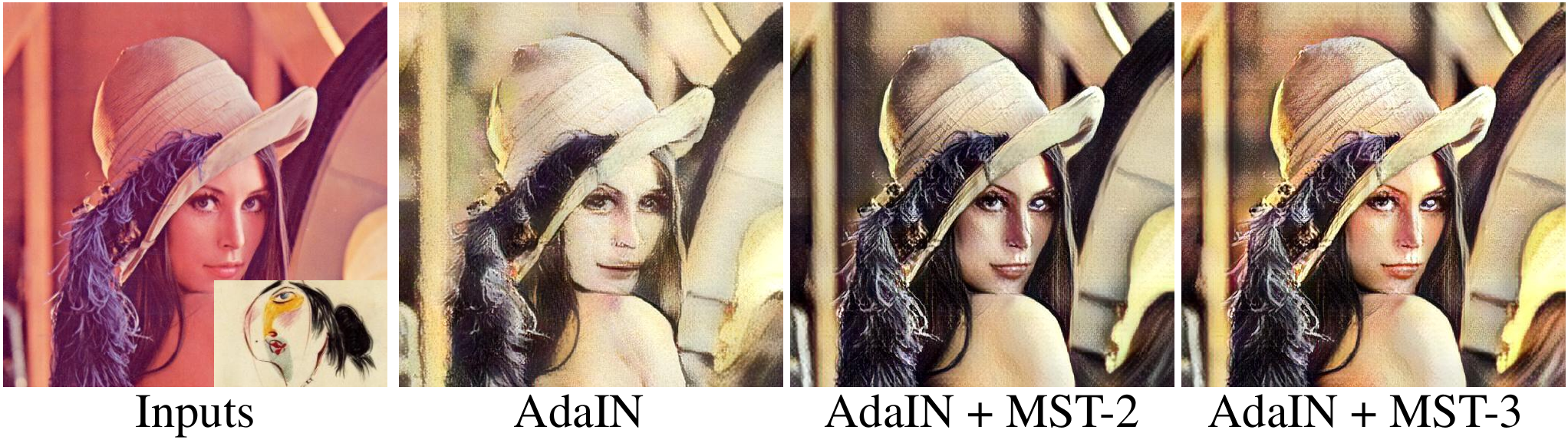}}
\vspace{-3mm}
\caption{Generalization of MST to AdaIN~\cite{huang2017arbitrary}.}
%Transferring features with WCT would alleviate the artifacts in the background and produce more natural stylizations than those with AdaIN.
\label{fig:mst_adain}
\vspace{-6mm}
\end{figure}
\vspace{-2mm}

\subsection{Generalization of MST}
\vspace{-2mm}
We further investigate the generalization of our proposed MST to improve some existing style transfer methods. Here, we take the popular AdaIN~\cite{huang2017arbitrary} as an example. We apply style clustering and graph based style matching to AdaIN, which is then denoted as ``AdaIN + MST-$K$''. As shown in Fig.~\ref{fig:mst_adain}, AdaIn may distort some content structures (e.g., mouth) by switching the global mean and standard deviation between style and content features. When we cluster the style feature into $K$ sub-sets and match them with content features via graph cuts, such a phenomenon can be obviously alleviated (see 3rd and 4th columns in Fig.~\ref{fig:mst_adain}). According to these observations and analyses, we can learn that our MST can be generalized and will benefit to some other existing style transfer methods.

%\subsection{Additional Experiments}

%\textbf{Photo Style Transfer}.

%\textbf{Video Stylization}.

%\textbf{Very High-Resolution Style Transfer}.

\vspace{-2mm}
\section{Conclusion}
\vspace{-2mm}
%We first propose multimodal style representation to model the complex style distribution. We then formulate the style matching problem as an energy minimization one and solve it using our proposed graph based style matching. As a result, we propose multimodal style transfer to transform features in a multimodal way. We not only treat the style patterns distinctively, but also consider the semantic content structure and its matching with style patterns. We also investigate that MST can be generalized to some existing style transfer methods and improve their stylization results. We conduct extensive experiments to validate the effectiveness, robustness, and flexibility of our method. \\
We propose multimodal style representation to model the complex style distribution. We then formulate the style matching problem as an energy minimization one and solve it using our proposed graph based style matching. As a result, we propose multimodal style transfer to transform features in a multimodal way. We treat the style patterns distinctively, and also consider the semantic content structure and its matching with style patterns. We also investigate that MST can be generalized to some existing style transfer methods. We conduct extensive experiments to validate the effectiveness, robustness, and flexibility of MST. \\
\textbf{Acknowledgements}: This work was supported by Adobe Research funding.

{\small
\bibliographystyle{ieee_fullname}
\bibliography{styletransfer_conf}
}

\end{document}